\documentclass[sigconf]{acmart}

\usepackage{booktabs}
\usepackage{multirow}
\usepackage{pifont}
\usepackage{bm}
\usepackage{algorithm}
\usepackage{algorithmicx}
\usepackage{algpseudocode}
\usepackage{amsmath}
\usepackage{makecell}
\usepackage{fancyhdr}
\usepackage{longtable}

\AtBeginDocument{%
  \providecommand\BibTeX{{%
    \normalfont B\kern-0.5em{\scshape i\kern-0.25em b}\kern-0.8em\TeX}}}

\setcopyright{acmcopyright}
\copyrightyear{2018}
\acmYear{2018}
\acmDOI{10.1145/1122445.1122456}

\acmConference[Woodstock '18]{Woodstock '18: ACM Symposium on Neural
  Gaze Detection}{June 03--05, 2018}{Woodstock, NY}
\acmBooktitle{Woodstock '18: ACM Symposium on Neural Gaze Detection,
  June 03--05, 2018, Woodstock, NY}
\acmPrice{15.00}
\acmISBN{978-1-4503-XXXX-X/18/06}

\begin{document}

\title{Are Negative Samples Necessary in Entity Alignment? An Approach with High Performance, Scalability and Robustness}

\author{Xin Mao$^{1*}$, Wenting Wang$^2$, Yuanbin Wu$^1$, Man Lan$^{1*}$}
\email{xmao@stu.ecnu.edu.cn, {wenting.wang}@lazada.com, {ybwu,mlan}@cs.ecnu.edu.cn}
\affiliation{$^1$East China Normal University \& $^2$Alibaba Group \country{China \& Singapore}}

\begin{abstract}
Entity alignment (EA) aims to find the equivalent entities in different KGs, which is a crucial step in integrating multiple KGs.
However, most existing EA methods have poor scalability and are unable to cope with large-scale datasets.
We summarize three issues leading to such high time-space complexity in existing EA methods:
(1) Inefficient graph encoders, (2) Dilemma of negative sampling, and (3) "Catastrophic forgetting" in semi-supervised learning.
To address these challenges, we propose a novel EA method with three new components to enable high \textbf{P}erformance, high \textbf{S}calability, and high \textbf{R}obustness (PSR):
(1) Simplified graph encoder with relational graph sampling, (2) Symmetric negative-free alignment loss,
and (3) Incremental semi-supervised learning.
Furthermore, we conduct detailed experiments on several public datasets to examine the effectiveness and efficiency of our proposed method.
The experimental results show that PSR not only surpasses the previous SOTA in performance but also has impressive scalability and robustness.
\end{abstract}



\maketitle

\section{Introduction}
\label{sec:intro}
Knowledge graphs (KGs) have facilitated many downstream applications, such as recommendation \cite{DBLP:conf/www/0003W0HC19,DBLP:conf/www/WangZZLXG19} and question-answering systems \cite{DBLP:conf/www/ZhaoXQB20,DBLP:conf/wsdm/QiuWJZ20}.
In KGs, the real-world facts are presented as triples $(h,r,t)$.
Over recent years, a large number of KGs have sprung up and been successfully applied to multiple fields.
These KGs are constructed from different data sources and languages by different organizations, so they usually hold unique information individually but also share an overlapping part in between.
Clearly, integrating them offers a broader view, which has been proven effective in some applications such as search engines and cross-border E-commerce.
For example, integrating cross-lingual KGs provides an opportunity to benefit the minority language users who usually suffer from lacking language resources.
Therefore, how to fuse knowledge from various KGs has attracted increasing attention.
\begin{figure}
    \centering
    \includegraphics[width=0.8\linewidth]{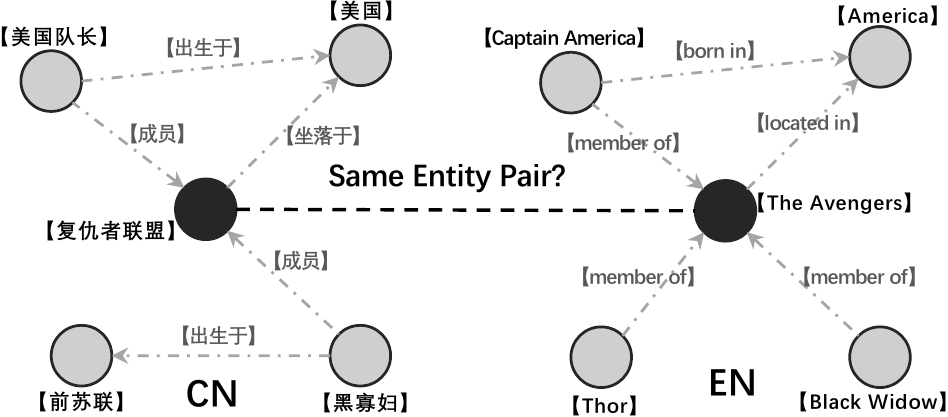}
    \caption{An example of cross-lingual entity alignment.}
    \label{fig:intro}
    \vspace{-1em}
\end{figure}

As shown in Figure \ref{fig:intro}, entity alignment (EA) aims to find the equivalent entities in different KGs, which is a crucial step in integrating multiple KGs.
With the incorporation of advanced techniques, the performances of EA methods are significantly improved over the past years. 
But on the other hand, the computation costs are also rapidly growing in terms of time and space complexity.
\citet{9174835} conduct an efficiency comparison of existing EA methods on a public dataset (DWY$100$K) that contains $100,000$ entity pairs and near one million triples.
In terms of time complexity, most advanced EA methods \cite{DBLP:conf/icml/GuoSH19,DBLP:conf/ijcai/SunHZQ18,DBLP:conf/acl/CaoLLLLC19,DBLP:conf/emnlp/YangZSLLS19} need to take more than $6$ hours for training and prediction, and several \cite{DBLP:conf/acl/XuWYFSWY19} even take days.
In terms of space complexity, several methods \cite{DBLP:conf/acl/CaoLLLLC19,DBLP:conf/ijcai/ZhuZ0TG19,DBLP:conf/emnlp/LiCHSLC19} require large memory consumption, so they are unable to run on a GPU with $12$ GB memory, even if setting with a minimum batch size.
Such poor time-space scalability hinders the application of existing EA methods to real-world datasets, which usually contain millions of entities and billions of triples (e.g., the full DBpedia contains $6.6+$ million entities, $23+$ billion triples).
We observe three issues below in existing EA methods and believe these are the causes of the above mentioned high time-space complexity bottleneck:

(1) \textbf{Inefficient graph encoders}:
Graph Neural Networks (GNNs) have become increasingly popular in addressing graph-based applications, including EA. 
The core of GNNs is that each node receives information from its adjacent nodes to update the structure-based embedding.
But with the expansion of receptive region (i.e., graph depth), the number of support nodes (and thus the time-space complexity) increases exponentially, a.k.a, "Neighbor Explosion."
Many graph sampling methods have been proposed to tackle this problem \cite{DBLP:conf/iclr/ChenMX18,DBLP:conf/kdd/ChiangLSLBH19,DBLP:conf/iclr/ZengZSKP20}.
However, these sampling methods only work on homogeneous graphs and ignore the types of edges, which usually capture important meta information.

In addition, many complicated techniques are adopted to improve performance, e.g., Graph Matching Networks \cite{DBLP:conf/icml/LiGDVK19} and Joint Learning \cite{DBLP:conf/acl/CaoLLLLC19}. 
The overall architectures of EA methods become more and more complex, while the time complexity is also dramatically increased.
For example, the running time of complex encoders (e.g., MuGNN \cite{DBLP:conf/acl/CaoLLLLC19}) is ten times more than that of the vanilla GCN \cite{DBLP:conf/emnlp/WangLLZ18}.

(2) \textbf{Dilemma of negative sampling}:
As a representation learning task, EA relies on margin-based pairwise loss functions (e.g., Triplet loss \cite{DBLP:conf/cvpr/SchroffKP15}, Contrastive loss \cite{DBLP:conf/cvpr/HadsellCL06}, or TransE \cite{DBLP:conf/nips/BordesUGWY13}).
In early studies \cite{DBLP:conf/semweb/SunHL17,DBLP:conf/emnlp/WangLLZ18}, negative samples are usually generated by uniform sampling and thus highly redundant.
During training stage, the model could be hampered by such low-quality negative samples, resulting in slow convergence and performance degradation.
Therefore, many previous studies focus on generating high-quality samples (i.e., hard samples), such as Top-$k$ loss \cite{DBLP:conf/nips/FanLYH17} and Focal loss \cite{DBLP:conf/iccv/LinGGHD17}.
In the EA task, BootEA \cite{DBLP:conf/ijcai/SunHZQ18} presents \emph{Truncated Uniform Negative Sampling Loss} to choose $k$-nearest neighbors as the hard negative samples.
Many subsequent studies \cite{DBLP:conf/acl/CaoLLLLC19,DBLP:conf/ijcai/ZhuZ0TG19} adopt this simple but effective strategy.
However, ranking all entities to find $k$-nearest neighbors in every epoch is extremely resource-consuming.
For instance, the sampling stage of BootEA takes up more than $25\%$ of the total time cost.
Moreover, a large number of negative samples implicitly link to a massive GPU memory requirement.
With the data scale further expanding from experimental datasets to real KGs, how to balance between performance and time-space consumption would become a dilemma.

(3) \textbf{"Catastrophic forgetting" in semi-supervised learning}:
"Catastrophic forgetting" refers to the phenomenon that the networks forget the previously learned samples when learning new samples.
In practice, it is expensive to label aligned entity pairs manually.
Therefore, existing EA studies \cite{DBLP:conf/emnlp/WangLLZ18,DBLP:conf/ijcai/SunHZQ18,DBLP:conf/ijcai/WuLF0Y019} usually reserve $30\%$ of the dataset or even less as the training data to simulate this situation.
Several EA methods \cite{DBLP:conf/ijcai/SunHZQ18,DBLP:conf/wsdm/MaoWXLW20} introduce iterative strategies to produce semi-supervised data, which significantly boosts up the performance.
However, due to the "catastrophic forgetting", existing semi-supervised EA methods have to mix up all the previously learned data with newly generated semi-supervised data and then re-train them altogether in the next iteration.
Thus, the running time of such iterative semi-supervised methods is usually several times more than that of a non-iterative method. 
With the expansion of the data scale, time consumption would continue to increase.

The above three challenges lead to the poor scalability of existing EA methods, rendering them infeasible for large-scale KGs.
In this paper, to address these challenges, we design the following three proposals:
(1) Remove inefficient components from existing EA methods to greatly simplify the graph encoder architecture and adopt a new graph sampling strategy based on relational attention mechanism.
(2) Inspired by recent contrastive image representation learning (CIRL) studies on the necessity of negative samples (BYOL \cite{DBLP:conf/nips/GrillSATRBDPGAP20} and SimSiam \cite{DBLP:journals/corr/abs-2011-10566}), we prove from another angle that the essence of GNNs-based EA methods is to solve a permutation matrix approximately and the negative samples are unnecessary in EA.
Based on this important finding, we design a symmetric negative-free alignment loss to improve the space and time efficiency significantly.
(3) To address the "catastrophic forgetting" phenomenon and further speed up the training speed, we demonstrate a new incremental semi-supervised learning strategy.
When learning newly generated semi-supervised samples, the model only needs to review a tiny amount of the previous samples but still learns effectively.
By incorporating the above three new components, we present a novel EA method with high \textbf{P}erformance, high \textbf{S}calability, and high \textbf{R}obustness (PSR).

To fully validate our proposed method, we conduct comprehensive experiments on several public datasets.
In terms of performance, the proposed method beats all state-of-the-art competitors across all datasets.
In ablation experiments, the performance of PSR only fluctuates by $3\%$, showing strong robustness on various hyper-parameter settings.
Most importantly, PSR demonstrates impressive scalability.
Its space complexity is only proportional to the batch size and graph density, while totally independent of the graph scale.
Meanwhile, its time complexity only increases linearly in accordance with the graph scale.
Under the same hardware environment condition, the speed of PSR is several times or even tens of times faster than other EA methods.
With a small batch size ($128$), our proposed method could run with only $3$ GB memory, while the performance is still comparable to SOTA (i.e., only degrading by less than $3\%$).
Our main contributions are summarized as follows:
\begin{itemize}
  \item To our best knowledge, this is the first work to prove that the essence of GNNs-based EA methods is to solve a permutation matrix approximately and explain why negative samples are unnecessary from a new angle of view.
  \item We propose a novel EA method with high \textbf{P}erformance, high \textbf{S}calability, and high \textbf{R}obustness (PSR), incorporating three components:
  (1) Simplified graph encoder with relational graph sampling. (2) Symmetric negative-free alignment loss. (3) Incremental semi-supervised learning.
  \item We design detailed experiments to examine the proposed method from multiple perspectives. The experimental results show that PSR not only surpasses the previous SOTA in performance but also has impressive scalability and robustness.

\end{itemize}

\section{Task Definition}
\label {sec:TF}
KGs store the real-world knowledge in the form of triples $(h,r,t)$.
A KG could be defined as $G=(E,R,T)$, where $E$ and $R$ represent the entity set and relation set respectively, $T\subset E\times R\times E$ denotes the triple set.
Defining $G_1$ and $G_2$ to be two KGs, $S$ is the set of pre-aligned entity pairs between $G_1$ and $G_2$. 
The task of EA aims to find new aligned pairs set $S'$ based on the pre-aligned seeds $S$.

\section{Related Work}
\subsection{Entity Alignment}
Most of existing EA methods can be summarized into two steps:
(1) Using KG embedding methods (e.g., TransE \cite{DBLP:conf/nips/BordesUGWY13}, GCN \cite{DBLP:journals/corr/KipfW16}, and GAT \cite{DBLP:conf/iclr/VelickovicCCRLB18}) to generate low-dimensional embeddings for entities and relations in each KG.
(2) Mapping these embeddings into a unified vector space through pre-aligned entities and pairing each entity by distance metrics (e.g., \textit{Cosine} and \textit{Manhattan}).
In addition, several methods apply iterative strategies \cite{DBLP:conf/ijcai/SunHZQ18,DBLP:conf/wsdm/MaoWXLW20} to generate semi-supervised data or introduce extra literal information \cite{DBLP:conf/acl/XuWYFSWY19,DBLP:conf/emnlp/WuLFWZ19,DBLP:conf/iclr/FeyL0MK20} (e.g., entity name) to enhance the model.
In Table \ref{tabel:rw}, EA methods are categorized based on what design is chosen for encoder and mapper and whether introducing the enhancement module.

\begin{table}
\begin{center}
\resizebox{1\linewidth}{!}{
\renewcommand\arraystretch{0.9}
\begin{tabular}{cccc}
  \toprule
  \textbf{Method}&\textbf{Encoder}&\textbf{Mapper}&\textbf{Enhancement}\\
  \toprule
  GCN-Align \cite{DBLP:conf/emnlp/WangLLZ18}&GNN&Margin&None\\
  MuGNN \cite{DBLP:conf/acl/CaoLLLLC19}&Hybrid&Margin&None\\
  RSNs \cite{DBLP:conf/semweb/SunHL17} &Trans&Corpus fusion&None\\
  HyperKA \cite{DBLP:conf/emnlp/SunCHWDZ20}&GNN&Margin&None\\
  \midrule
  BootEA \cite{DBLP:conf/ijcai/SunHZQ18}&Trans&Corpus fusion&Semi\\
  NAEA \cite{DBLP:conf/ijcai/ZhuZ0TG19}&Hybrid&Corpus fusion&Semi\\
  TransEdge\cite{DBLP:journals/corr/abs-2004-13579}&Trans&Corpus fusion&Semi\\
  MRAEA \cite{DBLP:conf/wsdm/MaoWXLW20}&GNN&Margin&Semi\\
  \midrule
  GM-Align \cite{DBLP:conf/acl/XuWYFSWY19}&GNN&Margin&Entity Name\\
  RDGCN \cite{DBLP:conf/ijcai/WuLF0Y019}&GNN&Margin&Entity Name\\
  HMAN \cite{DBLP:conf/emnlp/YangZSLLS19}&GNN& Margin&Attribute\\
  DGMC \cite{DBLP:conf/iclr/FeyL0MK20}&GNN&Margin&Entity Name\\
  \bottomrule
\end{tabular}
}
\end{center}
\caption{Categorization of some popular EA methods.}\label{tabel:rw}
\vspace{-1em}
\end{table}

\vspace{0.5em}
\noindent
\textbf{Graph Encoders}.
\emph{Trans} represents TransE \cite{DBLP:conf/nips/BordesUGWY13} and its derivative algorithms.
These methods interpret relations as a translation from head entities to tail entities and assume that the embeddings of entities and relations follow the assumption $h+r\approx t$.
Due to its easy implementation and high efficiency, the \emph{Trans} family is widely used in early EA methods.
In recent studies, GNNs gradually become the mainstream encoder because of their powerful graph modeling capability.
Furthermore, several EA methods adopt the \emph{Hybrid} strategy, i.e., using GNNs to model KGs and optimizing the \emph{Trans} loss at the same time.
However, \emph{Hybrid} methods usually require careful tuning on excessive hyper-parameters (e.g., NAEA has more than ten hyper-parameters), which leads to poor scalability and weak robustness.

\vspace{0.5em}
\noindent
\textbf{Mappers}.
\emph{Margin} indicates a series of margin-based pairwise losses, such as Triplet loss \cite{DBLP:conf/cvpr/SchroffKP15} and Contrastive loss \cite{DBLP:conf/cvpr/HadsellCL06}, which are often used in Siamese networks \cite{DBLP:conf/nips/BromleyGLSS93}.
Almost all GNNs-based EA methods adopt Siamese architectures with margin-based losses, which is quite similar to CIRL methods.
\emph{Corpus fusion} uses the pre-aligned set to swap the entities in existing triples and generates new triples to anchor the entities into a unified vector space.
For example, there are two triples $(e_1,r_1,e_2)\in KG_1$ and $(e_3,r_2,e_4)\in KG_2$.
If $(e_1,e_3) \in S$ holds, \emph{Corpus fusion} will add two extra triples $(e_3,r_1,e_2)$ and $(e_1,r_2,e_4)$.
\emph{Trans} methods usually apply such kind of mapper.

\vspace{0.5em}
\noindent
\textbf{Enhancement}.
Due to the lack of labeled data, several methods \cite{DBLP:conf/ijcai/SunHZQ18,DBLP:conf/ijcai/ZhuZ0TG19,DBLP:conf/wsdm/MaoWXLW20} adopt iterative strategies to construct semi-supervised data.
Despite significant performance improvements, the time consumption of these methods increases several times more.
Besides, some methods \cite{DBLP:conf/acl/XuWYFSWY19, DBLP:conf/ijcai/WuLF0Y019, DBLP:conf/emnlp/YangZSLLS19} include literal information (e.g., entity name) to provide a multi-aspect EA view.
However, literal information is not always available in real applications.
For example, there will be privacy risks when using user-generated content.

\subsection{Contrastive Image Representation Learning}

\begin{figure}
  \centering
  \includegraphics[width=1\linewidth]{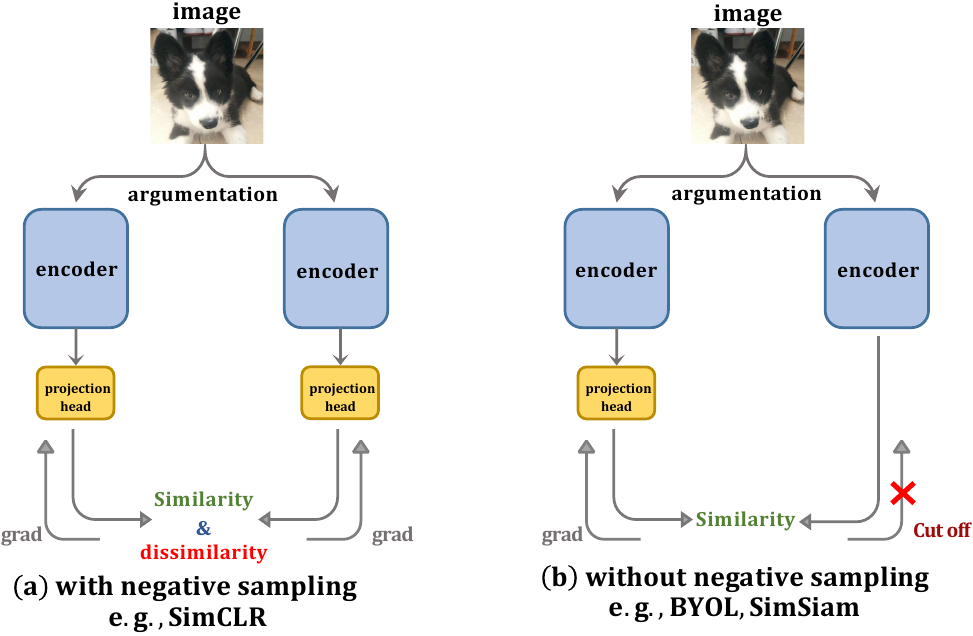}
  \caption{Examples of contrastive learning methods.\protect\footnotemark}
  \label{fig:rw}
\end{figure}
\footnotetext{His name is BianBian. What a lovely puppy, isn't it? }
The core idea of contrastive learning is to model samples through Siamese networks \cite{DBLP:conf/nips/BromleyGLSS93}, attract positive sample pairs, and repulse negative sample pairs.
This methodology has been widely utilized in self-supervised image representation learning.
Recent CIRL\cite{DBLP:conf/icml/ChenK0H20,DBLP:conf/nips/GrillSATRBDPGAP20,DBLP:journals/corr/abs-2011-10566} methods define the inputs as two augmentations of one image and maximize the similarity subject to different conditions (as shown in Figure \ref{fig:rw}).
The projection head is to reduce the loss of information induced by the contrastive losses. 
In practice, contrastive learning methods could benefit from a large number of negative samples, which require massive memory.
For example, SimCLR \cite{DBLP:conf/icml/ChenK0H20} regards all the other samples in the current batch as the negative samples, and the batch size is set to $8,192$ for best performance.
Obviously, most researchers can not afford the cost of such large-scale hardware resources.
MoCo \cite{DBLP:conf/cvpr/He0WXG20} presents the "momentum encoder" to alleviate this problem, but negative sampling remains to be the bottleneck.
More recently, BYOL \cite{DBLP:conf/nips/GrillSATRBDPGAP20} and SimSiam \cite{DBLP:journals/corr/abs-2011-10566} successfully prove that negative sampling is unnecessary for image contrastive learning by blocking half of the back-propagation stream (Figure \ref{fig:rw}b).

GNNs-based EA methods also rely on Siamese architecture with contrastive losses, so we believe that EA may also benefit from removing negative sampling, especially considering the space and time spent to find hard negative samples.

\section{Rethinking of Entity Alignment}
Recently, GNNs-based methods have become the mainstream of EA approaches.
Taking the very first and also the simplest GNNs-based EA method, GCN-align \cite{DBLP:conf/emnlp/WangLLZ18} as an example, it embeds each KG into a low-dimensional space via the following equation:
\begin{equation}
  \bm H^{(l+1)}=\sigma(\widetilde{\bm A}\bm H^{(l)}\bm W^{(l)})
  \label{eq1}
\end{equation}
where $\widetilde{\bm A} \in \mathbb{R}^{|E|\times |E|}$ is the symmetric normalized Laplacian matrix, $\sigma$ is an activation function, $\bm H^{(l)} \in \mathbb{R}^{|E|\times d}$ is the entity embedding matrix from the last layer, and $\bm W^{(l)}$ is a transformation matrix. 
Note that $\bm H^{(0)}$ is randomly initialized.
Then, these entity embeddings from two KGs are mapped into a unified space via Triplet loss:
\begin{equation}
  L = \sum_{({e_i},{e_j})\in S} \left[\gamma + \|\bm h_{e_i}-\bm h_{e_j}\| - \|\bm h_{e_i'}-\bm h_{e_j'}\|\right]_{+}
  \label{eq2}
\end{equation}
where $(e_i,e_j)$ is a pre-aligned pair, $(e_i',e_j')$ represents a negative pair generated by replacing one of $(e_i,e_j)$, $\gamma$ is the margin hyper-parameter, and $[x]_{+}$ indicates $max(x,0)$.
Of course, the actual implementation is more complicated, but the core idea can be summarized as the above.
Naturally, there is a question raised: \textit{What is implicitly optimized in GNNs-based EA methods?} 

\vspace{-1em}
\subsection{Another Perspective of EA}
\textbf{Our Hypothesis}: GNNs-based EA methods are implementations to solve a permutation matrix approximately.

\begin{figure}
  \centering
  \includegraphics[width=0.9\linewidth]{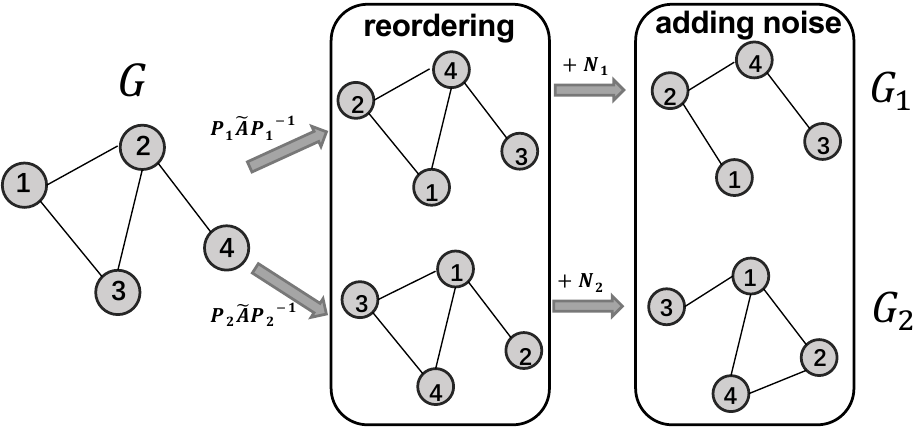}
  \caption{Construction process of $G_{1}$ and $G_{2}$ from $G$.}\label{fig:psample}
\end{figure}

Unlike the previous studies that regard EA as a representation learning task, we view EA from the perspective of permutation matrix solving.
Assuming there is an ideal graph $G$ containing all nodes and edges (i.e., the ground-truth graph), $G_1$ is just one particular graph instance constructed from $G$ by using information extraction algorithm or manual labor.
We use $\widetilde{\bm A} \in \mathbb{R}^{|E|\times |E|}$ and $\widetilde {\bm A}_1 \in \mathbb{R}^{|E|\times |E|}$ to denote the symmetric normalized Laplacian matrix of $G$ and $G_1$, respectively.
Then, the construction process of $G_1$ from $G$ could be described into two steps: reorder the entities in $G$ and introduce certain noise.
Formally, this construction process is formulated as below:
\begin{equation}
    \bm P\widetilde {\bm A}\bm P^{-1} + \bm N = \widetilde {\bm A}_1
    \label{eq3}
\end{equation}
where $\bm P \in \mathbb{R}^{|E|\times |E|}$ is a permutation matrix representing the equivalent relation of entities between $G$ and $G_1$.
Note that there is exactly one entry of $1$ in each row and each column in $\bm P$ while $0$s elsewhere.
Thus, $\bm P$ is also an orthogonal matrix, which complies with the fundamental assumption in EA that one entity has and only has one aligned entity.
When $\bm P_{ij}=1$, it represents that $e_i\in G$ and $e_j\in G_1$ are an aligned pair.
$\bm N \in \mathbb{R}^{|E|\times |E|}$ is a noise matrix denoting the subtle noise added during the construction process of $G_1$.
Thus, in essence, different graph instances could be regarded as the results of the same ideal graph by changing with different orders and introducing different noises (as illustrated in Figure \ref{fig:psample}).
Therefore, for two graph instances $G_1$ and $G_2$ constructed from the same ideal graph $G$, we have the following equation:
\begin{equation}
    {\bm P_1}^{-1}(\widetilde {\bm A_1}-\bm N_1)\bm P_1 = \widetilde {\bm A} = {\bm P_2}^{-1}(\widetilde {\bm A_2}-\bm N_2)\bm P_2
    \label{eq4}
\end{equation}
Let $\overline {\bm P}$ be the permutation matrix equaling to $\bm P_1{\bm P_2}^{-1}$. The above equation is transformed into:
\begin{equation}
    (\widetilde {\bm A_1}-\bm N_1)\overline{\bm P} = \overline{\bm P}(\widetilde {\bm A_2}-\bm N_2)
    \label{eq5}
\end{equation}
Meanwhile, according to the definition of the permutation matrix, the pre-aligned entity set $S$ is equivalent to a set of constraints:
\begin{equation}
    [(\widetilde {\bm A_1}-\bm N_1)\overline{\bm P}]_i = [(\widetilde {\bm A_2}-\bm N_2)]_j, \;\forall (e_i,e_j)\in S
    \label{eq6}
\end{equation}
where $[\bm X]_i$ represents the $i$-th row of matrix $\bm X$.
Obviously, the aim of EA now turns to solve the permutation matrix $\overline{\bm P}$ under these constraints (i.e., given pre-aligned entity pairs $S$).

\vspace{-0.5em}
\subsection{Approximate Matrix Solving}
\label{sec:AS}

Both the permutation matrix $\overline{\bm P}$ and the two noise matrices (i.e., $\bm N_{1}$ and $\bm N_{2}$) are unknown high-dimensional matrices, which is hard to solve directly. 
Without loss of generality, the EA task follows the two premises as below:

\textbf{Premise 1}:
Since the introduced noise is subtle, the structures of $G_1$ and $G_2$ are still isomorphic to $G$.
Thus, the noise matrices $\bm N_1$ and $\bm N_2$ in Equation (\ref{eq6}) could be dropped, approximately.

\textbf{Premise 2}: Approximately, the matrix $\overline{\bm P}$ could be further decomposed into the product of two low-rank dense matrices $\bm M_1 \in \mathbb{R}^{|E|\times d}$ and $\bm M_2 \in \mathbb{R}^{d\times |E|}$ (i.e., $\overline{\bm P} \approx \bm M_1 \bm M_2$).

Let ${\bm M_2}^{rinv}\in\mathbb{R}^{|E|\times d}$ be the right inverse matrix of $\bm M_2$ (i.e., $\bm M_2{\bm M_2}^{rinv}=I_d$), Equation (\ref{eq6}) could be transformed into:
\begin{equation}
    [\widetilde {\bm A_1}\bm M_1]_i = [\widetilde {\bm A_2}{\bm M_2}^{rinv}]_j, \;\forall (e_i,e_j)\in S
    \label{eq7}
\end{equation}
here $\bm M_1$ and ${\bm M_2}^{rinv}$ must be full column rank.
Obviously, when we modify the above equation by introducing an activation function $\sigma$ and adding further decompositions as $\bm M_1 = \bm H_1\bm W_1$ and ${\bm M_2}^{rinv} = \bm H_2\bm W_2$, Equation (\ref{eq7}) still holds:
\begin{equation}
    [\sigma(\widetilde {\bm A_1}\bm H_1\bm W_1)]_i = [\sigma(\widetilde {\bm A_2}\bm H_2\bm W_2)]_j, \;\forall (e_i,e_j)\in S
    \label{eq8}
\end{equation}
Compared to Equation (\ref{eq1}), we find that either side of Equation (\ref{eq8}) is in the same form as the right-hand side of Equation (\ref{eq1}), which suggests that the approximate matrix solving is equivalent to the Siamese networks with GCN encoders.
In general, Equation (\ref{eq8}) could be solved by the least-squares method:
\begin{equation}
    \underset{\bm H_{1,2},\bm W_{1,2}}{arg\;min}\;\sum_{(e_i,e_j)\in S} \left\|\sigma(\lbrack\widetilde{\bm A_1}\bm H_1\bm W_1\rbrack_i)\;-\;\sigma(\lbrack\widetilde{\bm A_2}\bm H_2\bm W_2\rbrack_j)\right\|_2^2
    \label{eq9}
\end{equation}
Simply optimizing Equation (\ref{eq9}) by gradient descent inclines to a trivial solution \cite{DBLP:journals/corr/abs-2011-10566} (i.e., all outputs collapse to a constant).
Therefore, it is essential to ensure that $\bm H_1\bm W_1$ and $\bm H_2\bm W_2$ are both full column rank to prevent the model from collapsing.

Generally, the involving of negative samples is equivalent to introducing a penalty term to trivial solutions, which can preclude constant outputs from the solution space.
Blocking half of the back-propagation stream (i.e., Figure \ref{fig:rw}b) also avoids constant outputs.
Putting this trick into our approximate solving framework, it is analogous to approximately solving $\overline{\bm P}$ by only updating $\bm{H}_1\bm{W}_1$ with a fixed $\bm{H}_2\bm{W}_2$.
Intuitively, in high dimensions, any randomly drawn vectors are always nearly orthogonal.
Thus, $\bm H_1\bm W_1$ and $\bm H_2\bm W_2$ are invariably full column rank.
In a nutshell, blocking back-propagation would guarantee that the EA models do not collapse.

\vspace{-0.5em}
\subsection{Further Analysis of GNNs-based EA}
\label{sec:dis}
In our opinion, GNNs-based EA has a close correlation with CIRL.
Once we view the ideal KG as the input image, the graph instance construction process is equivalent to the random image argumentation trick.
However, there are still several fundamental differences between these two tasks. For example, there is only one ideal graph in EA, but thousands of images in CIRL.
These differences might be the reason that several key techniques (e.g., momentum encoder) are proved to be effective in CIRL but not working in EA.


Note that although the above inferences depend on the premise that $\widetilde A$ is invariant, it is easily extended into attention-based GNNs \cite{DBLP:conf/iclr/VelickovicCCRLB18,DBLP:conf/ijcai/WuLF0Y019,DBLP:conf/wsdm/MaoWXLW20} with variant $\widetilde A$. 
Thereby a more general inference method is derived and proved in our subsequent experiments.


\section{The Proposed Method}
\label{model}
This section will describe our proposed method in detail, including the encoder architecture and training strategy.

\subsection{Encoder Architecture}
To effectively deal with large-scale KGs, we propose a minimalism relation-aware graph encoder.
The proposed encoder removes all high-complexity components (e.g., \textit{hybrid}  strategy and graph matching) and even abandons the normal linear transformation matrix.
The inputs of our model are two matrices: $\bm H^{(0)}_e \in \mathbb{R}^{|E|\times d}$ represents the input entity features and $\bm H_r \in \mathbb{R}^{|R|\times d}$ represents the input relation features.
Both of them are initialized by the random uniform initializer.
Our proposed minimalism relation-aware graph encoder consists of three major components:

\vspace{0.5em}
\noindent
\textbf{Relation-specific embeddings}.
KG usually contains complicated relationships, such as reflexive, one-to-many, many-to-one, and many-to-many.
Some KG embedding methods \cite{DBLP:conf/aaai/LinLSLZ15,DBLP:conf/esws/SchlichtkrullKB18} construct relation-specific entity embeddings in order to better describe these complex relations.
Inspired by RREA \cite{DBLP:conf/cikm/MaoWXWL20}, we generate reflections of entities along the relational hyperplanes to enforce relation-specificity:

\begin{equation}
  \phi(\bm h_{e_i},\bm h_{r_k}) = \bm h_{e_i} - 2\bm h_{r_k}^T\bm h_{e_i}\bm h_{r_k}
\end{equation}
where $\bm h_{e_i}$ and $\bm h_{r_k}$ represent the embeddings of entity $e_i$ and relation $r_k$, and $\|\bm h_{r_k}\|_2=1$.
Compared with RGCN \cite{DBLP:conf/aaai/LinLSLZ15} and TransR \cite{DBLP:conf/esws/SchlichtkrullKB18} which set up a transformation matrix for each relation, the above operation only requires much fewer parameters and has higher computational efficiency.

\vspace{0.5em}
\noindent
\textbf{Relational attention aggregator}.
When aggregating neighborhood information, the vanilla GCN only considers the degree of nodes and ignores the type of edges, which we believe also embeds hidden semantics.
Therefore, we bring in the relational attention mechanism to distinguish the importance between edges and aggregate neighboring information anisotropically.
Besides, we remove the normal linear transformation matrix and replace it with the relation-specific embeddings.
The output feature of $e_i$ from the $l$-$th$ layer is described as following:
\begin{equation}
\bm{h}_{e_i}^{(l+1)}= \sigma\left(\sum_{e_j\in \mathcal N_{e_i}}\sum_{r_k\in R_{ij}}\alpha_{ijk}^{(l)} \phi(\bm{h}_{e_j}^{(l)},\bm h_{r_k})\right)
\end{equation}
where $\mathcal N_{e_i}$ is the neighboring entity set of $e_i$, and $R_{ij}$ is the set of relations between $e_i$ and $e_j$.
In this work, we adopt ELU \cite{DBLP:journals/corr/ClevertUH15} as the activation function.
$\alpha_{ijk}^{(l)}$ represents the relational attention coefficient which is obtained as below:
\begin{equation}
  \beta_{ijk}^{(l)} = \bm{v}^T[\phi(\bm{h}_{e_i}^{(l)},\bm h_{r_k})\;\|\;\bm h_{r_k}\;\|\;\phi(\bm{h}_{e_j}^{(l)},\bm h_{r_k})]
\end{equation}
\begin{equation}
\alpha_{ijk}^{(l)}=\frac{exp(\beta_{ijk}^{(l)})}{\sum_{e_{j'}\in \mathcal N_{e_i}}\sum_{r_{k'}\in R_{ij'}}exp(\beta_{ij'k'}^{(l)})}
\end{equation}
where $\bm{v}\in \mathbb{R}^{3d}$ is a trainable vector for calculating the attention coefficient.
$\|$ means the concatenate operation.

\vspace{0.5em}
\noindent
\textbf{Multi-hop representation}.
To enable a global-aware representation, we concatenate the output embeddings from multiple layers, capturing multi-hop neighboring information.
The final output embedding of entity $e_i$ is in the following form:
\begin{equation}
 \bm{h}^{f}_{e_i} = [\bm{h}^{(0)}_{e_i}\|\bm{h}^{(1)}_{e_i}\| ...\| \bm{h}^{(l)}_{e_i}\ ]
\end{equation}
where $\bm{h}^{(0)}_{e_i}$ represents the input embedding of $e_i$.

\begin{figure}
  \centering
  \includegraphics[width=1\linewidth]{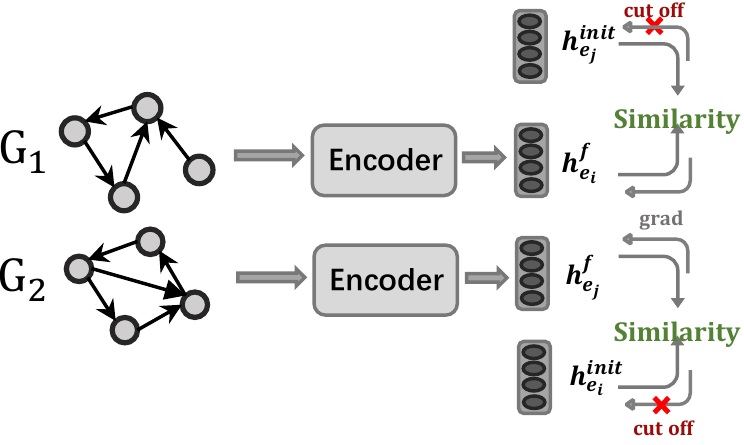}
  \caption{The illustration of symmetric negative-free loss.}\label{fig:negative-free}
\end{figure}

\begin{table*}[t]
\renewcommand\arraystretch{0.9}
\centering
\resizebox{0.9\textwidth}{!}{
\begin{tabular}{c|cc|cc|cc|cc|cc|cc|cc}
\hline
Datasets & $\rm{DBP_{ZH}}$ & $\rm{DBP_{EN}}$ & $\rm{DBP_{JA}}$ & $\rm{DBP_{EN}}$ & $\rm{DBP_{FR}}$ & $\rm{DBP_{EN}}$ & $\rm{SRP_{EN}}$ & $\rm{SRP_{FR}}$ & $\rm{SRP_{EN}}$ & $\rm{SRP_{DE}}$ & $\rm{DWY_{DBP}}$  & $\rm{DWY_{YG}}$ & $\rm{DWY_{DBP}}$  & $\rm{DWY_{WD}}$\\
\hline
$|E|$ & 19,388 & 19,572 & 19,814 & 19,780 & 19,661 & 19,993 & 15,000 & 15,000 & 15,000 & 15,000 & 100,000 & 100,000 & 100,000 & 100,000 \\
$|R|$ & 2,830 & 2,317 & 2,043 & 2,096 & 1,379 & 2,209 & 221 & 177 & 222 & 120 & 302 & 31 & 330 & 220 \\
$|T|$ & 70,414 & 95,142 & 77,214 & 93,484 & 105,998 & 115,722 & 36,508 & 33,532 & 38,363 & 37,377 & 428,952 & 502,563 & 463,294 & 448,774 \\
\hline
\end{tabular}
}
\caption{Statistics of the Datasets. $|E|$, $|R|$ and $|T|$ are the size of entities, relations and triples respectively.}
\label{dataset}
\vspace{-1em}
\end{table*}

\subsection{Training Strategy}
\label{sec:ts}
In addition to simplifying the encoder's architecture, the proposed method also boosts the training efficiency from three aspects:

\vspace{0.5em}
\noindent
\textbf{Relational attention graph sampling}.
Previous GNNs-based EA methods \cite{DBLP:conf/emnlp/WangLLZ18,DBLP:conf/wsdm/MaoWXLW20,DBLP:conf/acl/CaoLLLLC19} usually train on full graph, with the consequence of demanding large memory and slow speed.
Following GraphSAGE \cite{DBLP:conf/nips/HamiltonYL17}, we only sample a small batch of positive pairs from training data and construct a multi-hop sub-graph.
But instead of the equal probability sampling, we propose an unequal probability sampling strategy since we believe not all triples have equal importance.
Given the center node $e_i$, we sample $t$ times from its neighboring triples with replacement and directly use the normalized attention coefficient $\alpha_{ijk}$ as the probability of triple $(e_i,r_k,e_j)$:
\begin{equation}
  P((e_i,r_k,e_j)|e_i) = \alpha_{ijk}
\end{equation}
With this setting, more important triples have a higher probability of being sampled.
In accordance with this, for different center nodes, the expectations of attention coefficient from one sampling (also means information capacity) are different:
\begin{align}
  \mathbb{E}[\alpha_{e_i}] &= \sum_{e_j\in \mathcal N_{e_i}}\sum_{r_k\in R_{ij}} P((e_i,r_k,e_j)|e_i)\;\alpha_{ijk}\\
  &= \sum_{e_j\in \mathcal N_{e_i}}\sum_{r_k\in R_{ij}}{\alpha_{ijk}}^2
\end{align}

If we follow the previous studies to use a fixed $t$ or set $t$ inversely proportional to the degree, it will lead to redundancy or deficiency of information.
In fact, we expect that for any center node $e_i$, the sum of attention coefficient expectations should maintain certain stability after $t_{e_i}$ round of sampling:
\begin{equation}
  t_{e_i} = \lceil \frac{\tau}{\mathbb{E}[\alpha_{e_i}]} \rceil
\end{equation}
where $\tau$ is a hyper-parameter controlling the sampling ratio.
$\lceil x \rceil$ denotes the ceil operation.

\begin{figure}[t]
  \centering
  \includegraphics[width=0.90\linewidth]{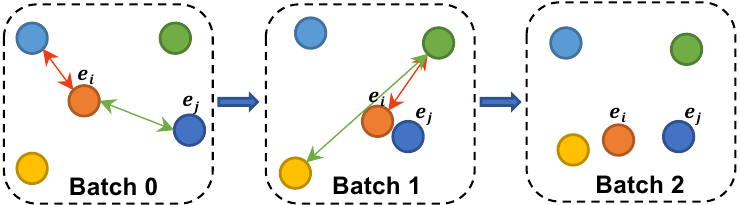}
  \caption{An example of "forgetting". Green lines represent positive pairs and red lines represent negative pairs.}\label{fig:forget}
  \vspace{-1em}
\end{figure}

\noindent
\textbf{Symmetric negative-free alignment loss}.
As mentioned in Section \ref{sec:AS}, one side of the Siamese networks' back-propagation should be blocked to prevent the model from collapsing in the case of no negative samples.
Specifically, we first initialize the graph encoder randomly, make a forward propagation on each KG, and store the initial output embedding $\bm h^{f}$ as $\bm h^{init}$ for each entity.
Next, taking $\bm h^{init}$ as the target, the model is optimized by the following loss function (as illustrated in Figure \ref{fig:negative-free}):
\begin{equation}
  L = \sum_{(e_i,e_j)\in S}-\; sim(\bm h_{e_i}^f,\bm h_{e_j}^{init})\;-\;sim(\bm h_{e_j}^f,\bm h_{e_i}^{init})
\end{equation}
The back-propagation streams of $\bm h_{e_i}^{init}$ and $\bm h_{e_j}^{init}$ are cut off.
In this paper, we use the \textit{cosine} similarity as the metric $sim(\bm x,\bm y)$.
\begin{equation}
  sim(\bm x,\bm y) = \frac{\langle \bm x,\bm y\rangle}{{\left\|\bm x\right\|}_2{\left\|\bm y\right\|}_2}
\end{equation}
In fact, $\bm h_{e_i}^{init}$ and $\bm h_{e_j}^{init}$ are analogous to the target network in BYOL \cite{DBLP:conf/nips/GrillSATRBDPGAP20}.
However, due to the fundamental differences between EA and CIRL, adopting the momentum encoder of MoCo and BYOL will cause the model to degrade.
Thus, the momentum encoder is not incorporated in our model. 
Once the training process is completed, the final similarity score between $e_i$ and $e_j$ is calculated as below:
\begin{equation}
  score(e_i,e_j) = sim(\bm h_{e_i}^f,\bm h_{e_j}^{init})\;+\;sim(\bm h_{e_j}^f,\bm h_{e_i}^{init})
\end{equation}

\vspace{0.5em}
\noindent
\textbf{Incremental semi-supervised learning}.
"Catastrophic forgetting" refers to the phenomenon that the networks forget the previously learned samples after learning new samples.
Specific in the EA task, $(e_i,e_j)\in S$ is a pre-aligned pair, which has already been mapped to an adjacent space in the early training batch.
However, once $e_i$ or $e_j$ is chosen to be negative samples, the model "forgets" their learned relative positions (as shown in Figure \ref{fig:forget}).
Therefore, existing EA methods need to mix the newly generated semi-supervised data with the early training data and then retrain the model, resulting in low efficiency.
With the complete absence of negative samples, the proposed method is born with the power to do incremental learning.
When learning new batches, the learned entity embeddings are only indirectly influenced by their neighboring entities.
Thus, the "forgetting" phenomenon is largely weakened.
To further alleviate "catastrophic forgetting," we store all the embeddings of the training set after each iteration (embeddings of the previous iteration will be overwritten).
In the next iteration, we compare them with the currently learned embeddings.
If the similarity difference of an entity pair exceeds $\epsilon$, such pair will be added into the next iteration again (a.k.a, reviewing).
According to our experiments, only very few pairs will be added multiple times.

In addition, we adopt the \emph{Bi-directional Iterative Strategy } proposed by MRAEA \cite{DBLP:conf/wsdm/MaoWXLW20} to construct high-quality semi-supervised aligned pairs.
If and only if the entities $e_i$ and $e_j$ are mutually nearest neighbors to each other, then the pair $(e_i,e_j)$ would be considered as new semi-supervised aligned entities.
The detailed incremental semi-supervised procedure is described in Algorithm \ref{algorithm}.

\renewcommand{\algorithmicrequire}{\textbf{Input:}}
\renewcommand{\algorithmicensure}{\textbf{Output:}}
\begin{algorithm}[t]
    \footnotesize
    \caption{Incremental Semi-supervised Learning Strategy}
    \begin{algorithmic}[1]
        \Require Graphs $G_1=(E_1,R_1,T_1), G_2=(E_2,R_2,T_2)$, pre-aligned seed entity pairs $S^{init}$. $E'_1\subseteq E_1, E'_2\subseteq E_2$ are the candidate entity sets not existing in $S^{init}$.
        \Ensure parameters $\theta$ for model
        \State Initialize the model;
        \State $S\leftarrow S^{init}$; \texttt{/*Initialize the training set.*/}
        \State $Q\leftarrow \{\}$; \texttt{/*Store the embeddings of all previous training pairs.*/}
        \Repeat
            \State Train the model with $G_1$, $G_2$, and $S$ until the loss of development set increase;
            \State $S^{new}\leftarrow \{\}$; \texttt{/*Initialize new training set.*/}
            \vspace{0.5em}
            \State \texttt{/*Review all the previous embeddings*/}
            \For {$(\bm h^{last}_{e_i},\bm h^{last}_{e_j}) \in Q$}
                \If {$sim(\bm h^{last}_{e_i},\bm h^{last}_{e_j})-sim(\bm h^{now}_{e_i},\bm h^{now}_{e_j})>\epsilon$}
                    \State $S^{new}\leftarrow S^{new}\cup \{(e_i,e_j)\}$;
                \EndIf
            \EndFor
            \State \texttt{/*Generate bi-directional semi-supervised aligned pairs*/}
            \For {$e \in E'_1$}
                \State $e'=N(e,E'_2)$ \texttt{/*$N(e,E)$ is the entity nearest to $e$ in $E$.*/}
                \If {$N(e',E'_1)=e$}
                    \State $S^{new}\leftarrow \left\{\left(e,e'\right)\right\}\cup S^{new}$;\;\;\;$E'_1\leftarrow E'_1-\{e\}$;\;\;\;$E'_2\leftarrow E'_2-\{e'\}$;
                \EndIf
            \EndFor
            \vspace{0.5em}
            \State \texttt{/*Update the embeddings set*/}
            \For {$(e_i,e_j)\in S$}
                    \state $Q\leftarrow Q\cup \{(\bm h^{now}_{e_i},\bm h^{now}_{e_j})\}$;
            \EndFor
            \State $S\leftarrow S^{new}$; \texttt{/*Update the training set*/}
        \Until{no more aligned entities are added to $S$}
    \end{algorithmic}\label{algorithm}
    \vspace{-0.5em}
\end{algorithm}

\begin{table*}[t]
\begin{center}
\resizebox{0.95\textwidth}{!}{
\renewcommand\arraystretch{0.85}
\begin{tabular}{c|cccccccccccccccc}
  \toprule
  \multicolumn{2}{c}{\multirow{2}{*}{Method}} & \multicolumn{3}{c}{$\rm{DBP_{ZH-EN}}$} & \multicolumn{3}{c}{$\rm{DBP_{JA-EN}}$} & \multicolumn{3}{c}{$\rm{DBP_{FR-EN}}$}& \multicolumn{3}{c}{$\rm{SRPRS_{FR-EN}}$}& \multicolumn{3}{c}{$\rm{SRPRS_{DE-EN}}$}  \\
  \multicolumn{2}{c}{} & H@1 & H@10 & MRR & H@1 & H@10 & MRR & H@1 & H@10 & MRR & H@1 & H@10 & MRR & H@1 & H@10 & MRR\\
  \hline
  \multirow{9}*{Basic}
  & MTransE & 0.209 & 0.512 & 0.310 & 0.250 & 0.572 & 0.360 & 0.247 & 0.577 & 0.360 &0.213& 0.447&0.290&0.107&0.248&0.160\\
  & GCN-Align & 0.434 & 0.762 & 0.550 & 0.427 & 0.762 & 0.540 & 0.411 & 0.772 & 0.530 & 0.243& 0.522&0.340&0.385&0.600& 0.460\\
  & MuGNN & 0.494 & 0.844 & 0.611 & 0.501 & 0.857 & 0.621 & 0.495 & 0.870  &0.621 &0.131&0.342& 0.208& 0.245&0.431&0.310\\
  & RSNs&0.508&0.745&0.591&0.507&0.737&0.590&0.516&0.768&0.605&0.350&0.636& 0.440& 0.484& 0.729&0.570\\
  & HyperKA&0.572&0.865&0.678& 0.564&0.865& 0.673&0.597&0.891& 0.704& -& -& -&- &-& -\\
  \cline{2-17}
  & \textbf{PSR(basic)} &\textbf{0.702}&\textbf{0.924}&\textbf{0.781}&\textbf{0.698}&\textbf{0.930}&\textbf{0.782}&\textbf{0.731}&\textbf{0.941}&\textbf{0.807}&\textbf{0.441}&\textbf{0.741}&\textbf{0.542}&\textbf{0.553}&\textbf{0.799}&\textbf{0.638}\\
  & $\rm{\pm stds}$&0.003&0.002&0.002&0.004&0.001&0.002&0.003&0.003&0.002&0.004&0.002&0.002&0.003&0.002&0.003\\
  & $p$-value &2e-16&8e-15&5e-17&2e-15&6e-18&3e-17&2e-16&1e-12&5e-17&8e-14&4e-17&6e-17&7e-14&2e-15&8e-14\\
  \hline
  \multirow{9}*{Semi}
  & BootEA & 0.629 & 0.847 & 0.703 & 0.622 & 0.853 & 0.701 & 0.653 & 0.874 & 0.731 & 0.365&0.649&0.460&0.503&0.732&0.580\\
  & NAEA & 0.650 & 0.867 & 0.720 & 0.641 & 0.872 & 0.718 & 0.673 & 0.894 & 0.752  &0.177&0.416& 0.260&0.307&0.535&0.390\\
  & TransEdge&0.735&0.919&0.801&0.719&0.932&0.795&0.710&0.941&0.796&0.400&0.675&0.490&0.556&0.753&0.630\\
  & JEANS &0.719&0.895&0.791&0.737&0.914&0.798&0.769&0.940&0.827&-&-&-&-&-&-\\
  & MRAEA &0.757&0.930&0.827&0.758&0.934&0.826&0.781&0.948&0.849&0.460&0.768&0.559&0.594&0.818&0.666\\
  \cline{2-17}
  & \textbf{PSR(semi)} &\textbf{0.802}&\textbf{0.935}&\textbf{0.851}&\textbf{0.803}&\textbf{0.938}&\textbf{0.852}&\textbf{0.828}&\textbf{0.952}&\textbf{0.874}&\textbf{0.486}&\textbf{0.781}&\textbf{0.577}&\textbf{0.606}&\textbf{0.831}&\textbf{0.680}\\
  & $\rm{\pm stds}$&0.004&0.001&0.004&0.003&0.002&0.003&0.003&0.002&0.002&0.003&0.002&0.003&0.004&0.003&0.002\\
  & $p$-value&4e-11&6e-08&1e-08&3e-12&1e-04&4e-10&2e-12&1e-04&2e-11&4e-10&6e-09&1e-08&4e-06&2e-07&3e-09\\
  \hline
  \multirow{8}*{Literal}
  & GM-Align & 0.679 & 0.785 & - & 0.739 & 0.872 & - & 0.894 & 0.952 & - & 0.574&0.646&0.602&0.681&0.748&0.710\\
  & RDGCN & 0.697 & 0.842 & 0.750 & 0.763 & 0.897 & 0.810 & 0.873 & 0.950 & 0.901  &0.672&0.767& 0.710&0.779&0.886&0.820 \\
  & HMAN&0.561&0.859&0.670&0.557&0.860&0.670&0.550&0.876&0.660&0.401&0.705&0.500&0.528&0.778&0.620\\
  & HGCN &0.720&0.857&0.760&0.766&0.897&0.810&0.892&0.961&0.910&0.670&0.770&0.710&0.763&0.863&0.801\\
  & DGMC &0.801&0.875&-&0.848&0.897&-&0.933&0.960&-&-&-&-&-&-&-\\
  \cline{2-17}
  & \textbf{PSR(lit)} &\textbf{0.883}&\textbf{0.982}&\textbf{0.928}&\textbf{0.908}&\textbf{0.987}&\textbf{0.939}&\textbf{0.958}&\textbf{0.997}&\textbf{0.975}&\textbf{0.808}&\textbf{0.933}&\textbf{0.853}&\textbf{0.881}&\textbf{0.970}&\textbf{0.914}\\
  & $\rm{\pm stds}$&0.004&0.002&0.003&0.005&0.003&0.002&0.003&0.001&0.001&0.003&0.004&0.002&0.003&0.002&0.003\\
  & $p$-value &2e-13&4e-17&2e-17&2e-11&7e-15&7e-18&6e-10&1e-15&6e-18&2e-16&4e-16&3e-18&2e-15&3e-16&4e-15\\
  \bottomrule
\end{tabular}
}
\caption{Experimental results ($\rm Means_{\pm stds}$) on DBP$15$K and SRPRS.
Besides the performances, we further conduct the one-sample T-test between PSR and the best baselines.
All the $p$-value $<0.01$ indicates that PSR significantly outperforms all baselines.}
\label{table:res1}
\vspace{-2em}
\end{center}
\end{table*}

\section{Experiments}
All the experiments are conducted on a PC with a GeForce GTX TITAN X GPU ($12$GB) and $128$GB memory.
The code is now available (\url{https://github.com/MaoXinn/PSR}).
\vspace{-0.5em}
\subsection{Datasets}
To fairly and comprehensively verify the performance, scalability, and robustness of our model, we experiment with three widely used public datasets:
(1) \textbf{DBP$15$K} \cite{DBLP:conf/semweb/SunHL17}:
This dataset consists of three cross-lingual subsets from multi-lingual DBpedia: $\rm DBP_{EN-FR}$, $\rm DBP_{EN-ZH}$, and $\rm DBP_{EN-JA}$.
Each subset contains $15,000$ pre-aligned entity pairs.
As an early dataset, DBP$15$K is the most popular one but has two defects: small scale and dense links.
(2) \textbf{SRPRS}:
\citet{DBLP:conf/icml/GuoSH19} propose this sparse dataset, including two cross-lingual subsets: $\rm SRPRS_{FR-EN}$ and $\rm SRPRS_{DE-EN}$.
Each subset also contains $15,000$ pre-aligned pairs but with much fewer triples compared to DBP$15$K .
(3) \textbf{DWY$100$K} \cite{DBLP:conf/ijcai/SunHZQ18}:
This dataset comprises two mono-lingual subsets, each containing $100,000$ pre-aligned entity pairs and nearly one million triples.
As the largest dataset, $\rm DWY100K$ raises challenges to the scalability of EA models.

The statistics of these datasets are summarized in Table \ref{dataset}.
To be consistent with previous studies \cite{DBLP:conf/emnlp/WangLLZ18,DBLP:conf/ijcai/SunHZQ18,DBLP:conf/ijcai/WuLF0Y019,DBLP:conf/wsdm/MaoWXLW20}, we randomly split $30\%$ of the pre-aligned entity pairs for training and development while using the remaining $70\%$ for testing.

\vspace{-1em}
\subsection{Baselines}
To fully evaluate our proposed method, we compare it against the following three groups of advanced EA methods:
(1) \textbf{Structure}:
These methods only use the original structure information (i.e., triples):
MTransE \cite{DBLP:conf/ijcai/ChenTYZ17}, GCN-Align \cite{DBLP:conf/emnlp/WangLLZ18}, RSNs \cite{DBLP:conf/icml/GuoSH19}, MuGNN \cite{DBLP:conf/acl/CaoLLLLC19}, HyperKA \cite{DBLP:conf/emnlp/SunCHWDZ20}.
(2) \textbf{Semi-supervised}:
These methods adopt iterative strategy to generate semi-supervised data:
BootEA \cite{DBLP:conf/ijcai/SunHZQ18}, NAEA \cite{DBLP:conf/ijcai/ZhuZ0TG19}, TransEdge \cite{DBLP:journals/corr/abs-2004-13579}, MRAEA \cite{DBLP:conf/wsdm/MaoWXLW20}, JEANS \cite{DBLP:conf/eacl/ChenSZR21}.
(3) \textbf{Literal}:
To obtain a multi-aspect view, these methods use the literal information (e.g., entity name) of entities as the input features:
GM-Align \cite{DBLP:conf/acl/XuWYFSWY19}, RDGCN \cite{DBLP:conf/ijcai/WuLF0Y019}, HMAN \cite{DBLP:conf/emnlp/YangZSLLS19}, HGCN \cite{DBLP:conf/emnlp/WuLFWZ19}, DGMC \cite{DBLP:conf/iclr/FeyL0MK20}.

To make a fair comparison against the above three groups of methods, PSR also has three corresponding versions:
(1) PSR (basic) is the basic version without incremental learning.
(2) PSR (semi) introduces incremental learning to generate semi-supervised data.
(3) PSR (lit) adopts a simple strategy to incorporate the literal information.
Specifically, for $e_i$ and $e_j$, we first use PSR (Semi) to obtain the structural similarity $ss_{ij}$.
Then, using the cross-lingual word embeddings (same with GM-Align \cite{DBLP:conf/iclr/LampleCRDJ18}) to calculate the literal similarity $ls_{ij}$.
Finally, the entities are ranked according to $ls_{ij} + ss_{ij}$.

\vspace{-1em}
\subsection{Settings}
\textbf{Metrics}.
Following convention \cite{DBLP:conf/ijcai/ChenTYZ17,DBLP:conf/emnlp/WangLLZ18,DBLP:conf/ijcai/SunHZQ18,DBLP:conf/acl/XuWYFSWY19}, we use $Hits@k$ and \emph{Mean Reciprocal Rank} ($MRR$) as the evaluation metrics.
We report the average of five independent runs as the results.

\noindent
\textbf{Hyper-parameters}.
For all datasets, the same default configuration is set:
embedding dimension $d = 300$;
depth of GNN $l = 2$;
sampling ratio $\tau=1$;
reviewing threshold $\epsilon = 0.05$;
batch size is $512$; dropout rate is set to $30\%$.
RMSprop is adopted to optimize the model with a learning rate set to $0.005$.

\vspace{-1em}
\subsection{Main Experiments}
In Table \ref{table:res1} and Table \ref{table:res2}, we report the performance of all methods on all three datasets.
Our method consistently achieves the best performance across all datasets and metrics.

\noindent
\textbf{Results on DBP$\bm 15$K}.
On this dataset, PSR (basic) outperforms the previous SOTA in the Basic group by more than $13\%$ on Hit@1 and $10\%$ on MRR.
This indicates that the architecture of PSR effectively captures the rich relational structure information existing in this dataset.
Benefiting from the extra semi-supervised data, the performance of PSR (semi) is significantly improved on every metric compared to its basic version. 
Moreover, compared against other advanced semi-supervised EA methods, PSR (semi) still maintains a significant performance advantage of $4\%$ on $Hits@1$.
Through further incorporating of cross-lingual word embeddings, PSR (lit) could model entities from both structure and semantics perspectives.
Therefore, its performance further increases and consistently surpasses the previous SOTA in the Literal group.
Besides, we observe that the performances of the Literal group vary significantly across language pairs, which is entirely different from the structure-only groups.
The French dataset has the most apparent gain, while the Chinese dataset has the least.

\noindent
\textbf{Results on SRPRS}.
To simulate the degree distribution in real-world data, SRPRS greatly cuts down the number of triples, which challenges EA methods' embedding capability on sparse KGs.
Although the performance of PSR drops compared to DBP$15$K, it still outperforms all the other advanced EA methods.
The semi-supervised strategy still has some benefits, but the improvement is reduced to $4$-$5\%$.
We believe the reason for this smaller improvement is that the sparse nature of SRPRS makes the generation of high-quality semi-supervised data very hard.
Therefore, the literal information becomes critical on SRPRS and improves $Hits@1$ by at least $28\%$ if compared to PSR (semi).

\begin{table}[t]
\begin{center}
\resizebox{1\linewidth}{!}{
\renewcommand\arraystretch{1}
\begin{tabular}{c|cccccccccccccccc}
  \toprule
  \multicolumn{2}{c}{\multirow{2}{*}{Method}} & \multicolumn{3}{c}{$\rm{DWY_{DBP-WD}}$}& \multicolumn{3}{c}{$\rm{DWY_{DBP-YG}}$}  \\
  \multicolumn{2}{c}{} & H@1 & H@10 & MRR & H@1 & H@10 & MRR\\
  \hline
  \multirow{8}*{Basic}
  & MTransE &0.238&0.507&0.330&0.227&0.414&0.290\\
  & GCN-Align&0.494&0.756&0.590&0.598&0.829&0.680\\
  & MuGNN &0.604&0.894&0.701&0.739&0.937&0.810\\
  & RSNs&0.607&0.793&0.673&0.689&0.878&0.756\\
  \cline{2-8}
  & \textbf{PSR(basic)}&\textbf{0.781}&\textbf{0.935}&\textbf{0.838}&\textbf{0.851}&\textbf{0.968}&\textbf{0.894}\\
  & $\rm{\pm stds}$&0.003&0.001&0.002&0.003&0.003&0.002\\
  & $p$-value&1e-17&4e-16&4e-18&9e-16&9e-11&3e-16\\
  \hline
  \multirow{7}*{Semi}
  & BootEA &0.748&0.898&0.801&0.761&0.894&0.808\\
  & NAEA &0.767&0.918&0.817&0.779&0.913&0.821\\
  & TransEdge&0.788&0.938&0.824&0.792&0.936&0.832\\
  & MRAEA&0.794&0.930&0.856&0.819&0.951&0.875\\
  \cline{2-8}
  & \textbf{PSR(semi)}&\textbf{0.881}&\textbf{0.967}&\textbf{0.912}&\textbf{0.892}&\textbf{0.978}&\textbf{0.923}\\
  & $\rm{\pm stds}$&0.002&0.001&0.002&0.003&0.001&0.002\\
  & $p$-value&2e-16&1e-15&1e-14&4e-14&2e-14&5e-14\\
  \bottomrule
\end{tabular}
}
\caption{Experimental results on DWY$100$K\protect\footnotemark.}
\label{table:res2}
\end{center}
\vspace{-1.5em}
\end{table}
\footnotetext{According to \citet{9174835}, entity names between mono-lingual KGs for this dataset are almost identical and the edit distance algorithm could achieve the ground-truth performance.
Therefore, the experimental results of literal methods are not listed.}

\begin{table}[t]
\renewcommand\arraystretch{1}
\centering
\resizebox{1.0\linewidth}{!}{
\begin{tabular}{l|cccccc}
\toprule
\multirow{2}{* }{Method} & \multicolumn{2}{c}{$\rm{DBP_{ZH-EN}}$} & \multicolumn{2}{c}{$\rm{DBP_{JA-EN}}$} & \multicolumn{2}{c}{$\rm{DBP_{FR-EN}}$}\\
& Hits@1 & MRR & Hits@1 & MRR & Hits@1 & MRR\\
\toprule
 PSR(semi) &$0.802$&$0.851$&$0.803$&$0.852$&$0.828$&$ 0.874$\\
 \;\;-RSE.&$0.790$&$0.841$&$0.774$&$0.828$&$0.806$&$ 0.854$\\
 \;\;-RAA.&$0.789$&$0.842$&$0.785$&$0.841$&$0.809$&$ 0.862$\\
 \;\;-MHR. &$0.751$&$0.817$&$0.750$&$0.815$&$0.789$&$ 0.847$\\
 \;\;-RAGS. &$0.789$&$0.843$&$0.788$&$0.844$&$0.816$&$ 0.868$\\
 \;\;-SNAL. &$0.799$&$0.850$&$0.798$&$0.847$&$0.830$&$ 0.876$\\
\bottomrule
\end{tabular}
}
\caption{Ablation experiments of model architecture.}
\label{table:model_ablation}
\vspace{-2em}
\end{table}


\noindent
\textbf{Results on DWY$\bm 100$K}.
As the largest dataset, DWY$100$K raises challenges to the space-time complexity of EA methods.
MuGNN and NAEA exceed the GPU memory limit setup in this experiment, so they have to run on the CPU, which substantially slows down the training speed.
On the contrary, the high scalability enables our model easy to cope with this large dataset and surpasses the previous SOTA by at least $7\%$ on $Hits@1$ (shown in Table \ref{table:res2}).
In addition, since this dataset shares similar dense degree distribution with DBP15K, the semi-supervised strategy also works well here.

\noindent
\textbf{Time Efficiency}.
The trump card of PSR is superior efficiency and scalability.
So we specifically evaluate the overall time costs of existing EA methods and report all results in Table \ref{tabel:time}.
It is obvious that the efficiency of PSR far exceeds all advanced competitors.
The speed of our proposed method is tens or even hundreds of times faster than that of existing complex EA methods.
Even compared against the fastest baseline (i.e., GCN-Align), the speed of PSR (basic) is $5\times$ faster on DWY$100$K, while the $Hits@1$ is $25\%$ higher.
Among the PSR group itself, the semi-supervised strategy consumes $2\times$ more time cost than the basic version, while PSR (lit) has no clear additional time cost than PSR (semi).


\begin{table}[t]
\begin{center}
\resizebox{0.9\linewidth}{!}{
\renewcommand\arraystretch{0.85}
\begin{tabular}{c|cccc}
  \toprule
  \multicolumn{2}{c}{Method}&\textbf{DBP15K}&\textbf{SRPRS}&\textbf{DWY100K}\\
  \toprule
  \multirow{6}*{Basic}
  &MTransE \cite{DBLP:conf/ijcai/ChenTYZ17} &6,467&3,355&70,085\\
  &GCN-Align \cite{DBLP:conf/emnlp/WangLLZ18}&103&87&3,212\\
  &RSNs \cite{DBLP:conf/icml/GuoSH19}&7,539&2,602&28,516\\
  &MuGNN \cite{DBLP:conf/acl/CaoLLLLC19}&3,156&2,215&47,735\\
  &\textbf{PSR(basic)} \cite{DBLP:conf/aaai/SunW0CDZQ20}&\textbf{88}&\textbf{75}&\textbf{603}\\
  \midrule
  \multirow{5}*{Semi}
  &BootEA \cite{DBLP:conf/ijcai/SunHZQ18}&4,661&2,659&64,471\\
  &NAEA \cite{DBLP:conf/ijcai/ZhuZ0TG19}&19,115&11,746&171,357\\
  &TransEdge\cite{DBLP:journals/corr/abs-2004-13579}&3,629&1,210&20,839\\
  &MRAEA \cite{DBLP:conf/wsdm/MaoWXLW20}&3,894&1,248&23,275\\
  &\textbf{PSR(semi)} \cite{DBLP:conf/aaai/SunW0CDZQ20}&\textbf{186}&\textbf{168}&\textbf{1,452}\\
  \midrule
  \multirow{5}*{Literal}
  &GM-Align \cite{DBLP:conf/acl/XuWYFSWY19}&26,328&13,032&459,715\\
  &RDGCN \cite{DBLP:conf/ijcai/WuLF0Y019}&6,711&886&-\\
  &HMAN \cite{DBLP:conf/emnlp/YangZSLLS19}&5,455&4,424&31,895\\
  &HGCN \cite{DBLP:conf/emnlp/WuLFWZ19}&11,275&2,504&60,005\\
  &\textbf{PSR(lit)} \cite{DBLP:conf/aaai/SunW0CDZQ20}&\textbf{195}&\textbf{183}&\textbf{1,532}\\
  \bottomrule
\end{tabular}
}
\end{center}
\caption{Time costs of EA methods (seconds).\protect\footnotemark}\label{tabel:time}
\vspace{-1em}
\end{table}
\footnotetext{All results are obtained by directly running the source code with default settings.
RDGCN requires extremely high memory space and we are unable to obtain its result on DWY$100$K.}

\subsection{Ablation Experiment}

\begin{figure}
  \centering
  \includegraphics[width=1\linewidth]{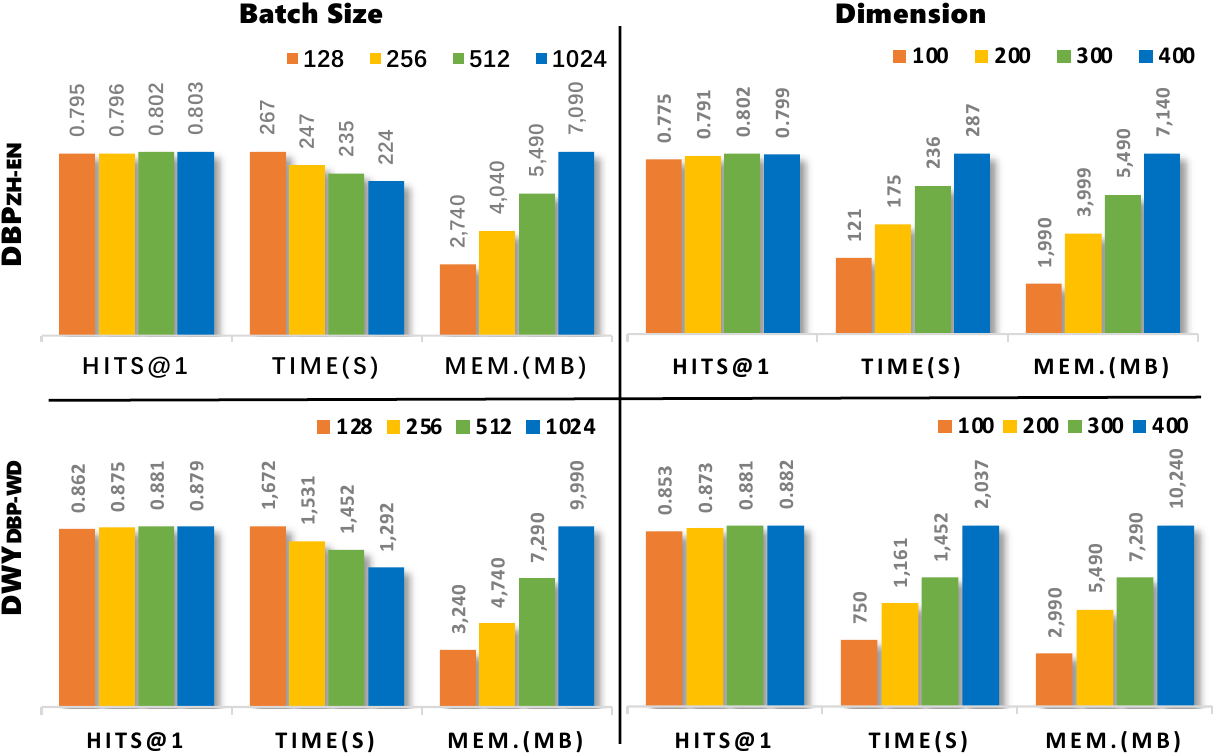}
  \caption{Ablation experiments on batch size and dimension.}\label{fig:chart}
\end{figure}

\noindent
\textbf{Model Architecture}.
The architecture of PSR consists of the following five components:
(1) \emph{Relation-specific embeddings} (RSE);
(2) \emph{Relational attention aggregator} (RAA);
(3) \emph{Multi-hop representation} (MHR);
(4) \emph{Relational attention graph sampling} (RAGS);
(5) \emph{Symmetric negative-free alignment loss} (SNAL).
We replace these components from PSR(semi) individually to demonstrate their effectiveness.
Specifically, RSE is replaced by normal linear transformation matrices.
RAA and RAGS become isotropic, i.e., all entities and relations have equal importance.
MHR is replaced by the residual connection \cite{DBLP:conf/cvpr/HeZRS16} and SNAL is replaced by the \emph{Truncated Uniform Negative Sampling loss} \cite{DBLP:conf/ijcai/SunHZQ18}.
The experimental results are listed in Table \ref{table:model_ablation}.
Among all these components, MHR has the greatest impact on performance.
Without MHR, the performances are degraded by at least $4\%$ on $Hits@1$.
For SNAL, the results are almost identical to the \emph{Truncated Uniform Negative Sampling loss}, indicating that SNAL could reduce the time-space consumption without losing any accuracy.
Besides, the remaining three components also show the necessity as our expectation.
On average, adopting them improves performance by $1\%$ to $3\%$ on $Hits@1$.
\clearpage

\noindent
\textbf{Batch Size}.
To explore the impact of batch size, we report the performances of PSR (semi) with batch size from $128$ to $1,024$ on $\rm DBP_{ZH-EN}$ and $\rm DWY_{DBP-WD}$.
In this experiment, only the batch size changes, all the other hyper-parameters remain fixed.
As shown in Figure \ref{fig:chart}, our method constantly works well over this wide range of batch sizes on both datasets.
Even a small batch size of $128$ performs decently, with a drop of less than $2\%$ on $Hits@1$.
We observe that the batch size greatly impacts the GPU memory occupation almost linearly.
Since PSR adopts a mini-batch training strategy, its space complexity is expected to be only related to batch size and graph density, not to the scale of graphs. 
The small memory consumption gap between DBP$15$K and DWY$100$K indeed verify this expectation. 

\noindent
\textbf{Embedding Dimension}.
Figure \ref{fig:chart} also reports the performances with embedding dimension $d$ from $100$ to $400$.
Similar to batch size, PSR also works well over this wide range of dimensions. 
Even a dimension size of $100$ performs decently, with a drop of only $3\%$ on $Hits@1$.
It is very obvious that both the time and space costs decrease linearly with the shrink of embedding dimension.
Overall, PSR is capable of dealing with large-scale datasets and run on devices with limited memory but still output comparable performance.

\noindent
\textbf{Incremental Learning}.
In Section \ref{sec:ts}, we mention that PSR is born with incremental learning ability and propose a simple reviewing strategy to further alleviate "catastrophic forgetting".
To validate the effectiveness of our design, we set the reviewing threshold $\epsilon$ from $0.01$ to $0.1$.
Furthermore, there are two baselines:
(a) \emph{Normal} represents existing EA methods' semi-supervised strategy, i.e., mixing all the newly generated data and early data into the next iteration.
(b) \emph{w/o rev.} means not reviewing any early samples at all (i.e., $\epsilon=\infty$).
In this way, each pair will be added to the training set only once.

\begin{figure}
  \centering
  \includegraphics[width=1\linewidth]{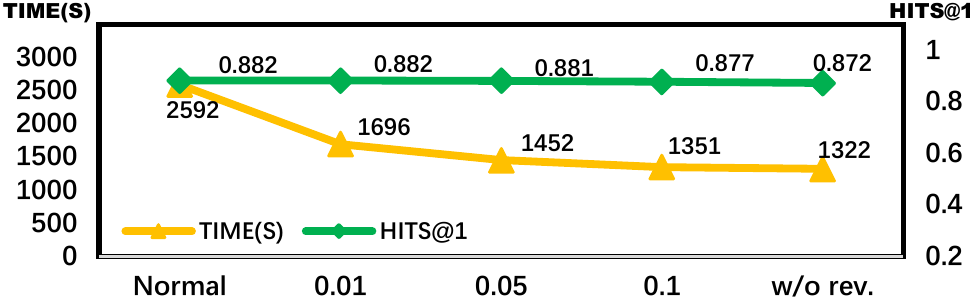}
  \caption{$\bm{Hits@1}$ and time cost of PSR (semi) with different reviewing threshold $\epsilon$ on $\rm {DWY_{DBP-WD}}$.}\label{fig:review}
\end{figure}

As shown in Figure \ref{fig:review}, the performance gap between these two baselines is only $1\%$, which proves that PSR is capable of learning incrementally.
By further incorporating the complete reviewing mechanism, PSR could significantly reduce the time cost while keeping the performance intact.
Such an incremental learning strategy could ensure that the training complexity increases linearly with the dataset scale, making massive-scale EA feasible.

\noindent

\begin{figure}
  \centering
  \includegraphics[width=\linewidth]{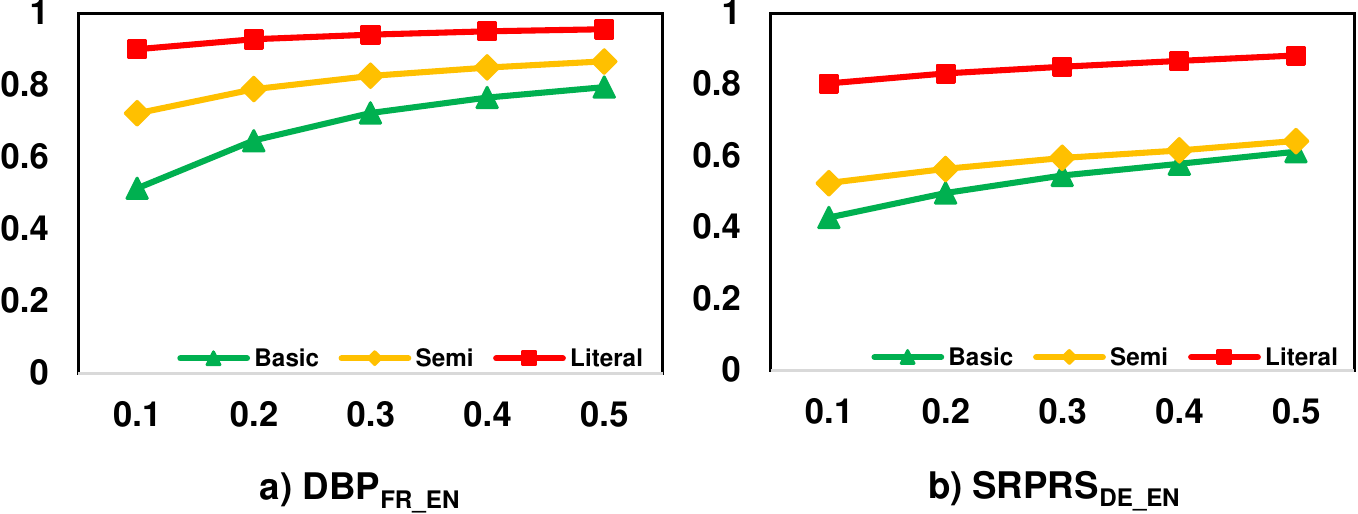}
  \caption{$\bm{Hits@1}$ of PSR with different pre-aligned ratios.}\label{fig:seed}
\end{figure}

\noindent
\textbf{Pre-aligned Ratio}.
In practice, manually annotating pre-aligned entity pairs consumes a lot of resources, especially for large-scale KGs.
We expect that the proposed model could keep decent performance with limited pre-aligned entity pairs.
To investigate the performance with different pre-aligned ratios, we set the ratios from $10\%$ to $50\%$.
Figure \ref{fig:seed} shows the results on $\rm DBP_{FR-EN}$ and $\rm SRPRS_{DE-EN}$.
Benefiting from the rich structure information in $\rm DBP_{FR-EN}$, the model could iteratively generate a large number of high-quality entity pairs for training.
Therefore, the semi-supervised strategy performs pretty well and greatly improves the performance.
On the other hand, because $\rm SRPRS_{DE-EN}$ is much more sparse, the impact of the semi-supervised strategy is very limited.
In both datasets, introducing literal information could significantly improve the performance.
Especially in $\rm SRPRS_{DE-EN}$, literal information could complement the insufficiency of structural information.

\begin{figure}
  \centering
  \includegraphics[width=1\linewidth]{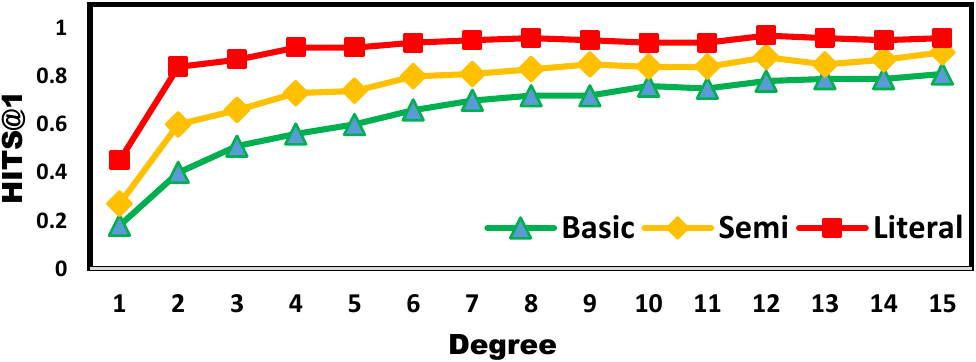}
  \caption{$\bm{Hits@1}$ of the entities with different degrees.}\label{Figure:error}
\end{figure}

\noindent
\textbf{Degree Analysis}.
The above experiments show that the performances of all EA methods on sparse datasets is much worse than that of standard datasets.
To further explore the correlation between performance and density, we design an experiment on $\rm DBP_{FR-EN}$.
Figure \ref{Figure:error} shows the $Hits@1$ of the three variants on different levels of entity degrees.
We observe that there is a strong correlation between performance and degree.
The semi-supervised strategy does improve the overall performances but has a limited effect on entities whose local structures are extremely sparse.
On the other hand, PSR (lit) has much better performances.
Unfortunately, literal information is not always available in real-life applications.
Therefore, how to better represent these sparse entities without extra information could be one key point for future work.

\section{Conclusion}
In this paper, we prove that the essence of GNNs-based EA methods is to solve a permutation matrix approximately and explain why negative samples are
unnecessary from a new angle of view.
Furthermore, we propose a novel EA method with three major modifications: (1) Simplified graph encoder with relational graph sampling. (2) Symmetric alignment loss without negative sampling. (3) Incremental semi-supervised learning strategy. 
Thus, the proposed method could simultaneously possess high performance, high scalability, and high robustness.
\bibliographystyle{ACM-Reference-Format}
\bibliography{sample-base}

\end{document}